\documentclass[10pt,twocolumn,letterpaper]{article}

 \usepackage[pagenumbers]{iccv} 

\definecolor{iccvblue}{rgb}{0.21,0.49,0.74}
\usepackage[pagebackref,breaklinks,colorlinks,allcolors=iccvblue]{hyperref}

\def\paperID{3845} 
\def\confName{ICCV}
\def\confYear{2025}

\title{AnimatePainter: A Self-Supervised Rendering Framework for Reconstructing Painting Process}

\author{
 Junjie Hu\textsuperscript{\rm 1}, Shuyong Gao\textsuperscript{\rm 1,2,*},  Qianyu Guo\textsuperscript{\rm 1},\\
      Yan Wang\textsuperscript{\rm 1}, Qishan Wang\textsuperscript{\rm 1},
      Yu'ang Feng\textsuperscript{\rm 1}, Wenqiang Zhang\textsuperscript{\rm 1,*} \\ 
    \textsuperscript{\rm 1} Fudan University, Shanghai, China \\
    \textsuperscript{\rm 2} Keenon Robotics Co. Ltd, Shanghai, China
    }

\begin{document}
\maketitle
\begin{abstract}
%

Humans can intuitively decompose an image into a sequence of strokes to create a painting, yet existing methods for generating drawing processes are limited to specific data types and often rely on expensive human-annotated datasets. We propose a novel self-supervised framework for generating drawing processes from any type of image, treating the task as a video generation problem. Our approach reverses the drawing process by progressively removing strokes from a reference image, simulating a human-like creation sequence. Crucially, our method does not require costly datasets of real human drawing processes; instead, we leverage depth estimation and stroke rendering to construct a self-supervised dataset. We model human drawings as "refinement" and "layering" processes and introduce depth fusion layers to enable video generation models to learn and replicate human drawing behavior. Extensive experiments validate the effectiveness of our approach, demonstrating its ability to generate realistic drawings without the need for real drawing process data.

\end{abstract}    
\section{Introduction}
\label{sec:introduction}

Painting is a manifestation of human artistry, which can also be an AI ability. We already have Diffusion \cite{stable-diffusion} and GAN \cite{gan} to generate impressive images that rival or even surpass humans. However, a complete image is only enough for us to appreciate, and cannot provide us with painting guidance. Painters need to deconstruct a painting into process fragments to learn and improve their painting skills, while painting robots \cite{frida,co-frida,robotPaint1} need intuitive fragments to draw vivid paintings.
While limited to certain types of painting, some methods \cite{paintsundo,InversePainting,timecraft} can accurately reconstruct the painting process of a painting. Most of them train on real drawing data which is expensive and difficult to obtain. Specific and scarce training data limits the types of images they can generate. 

\begin{figure}
    \centering
    \includegraphics[width=1\linewidth]{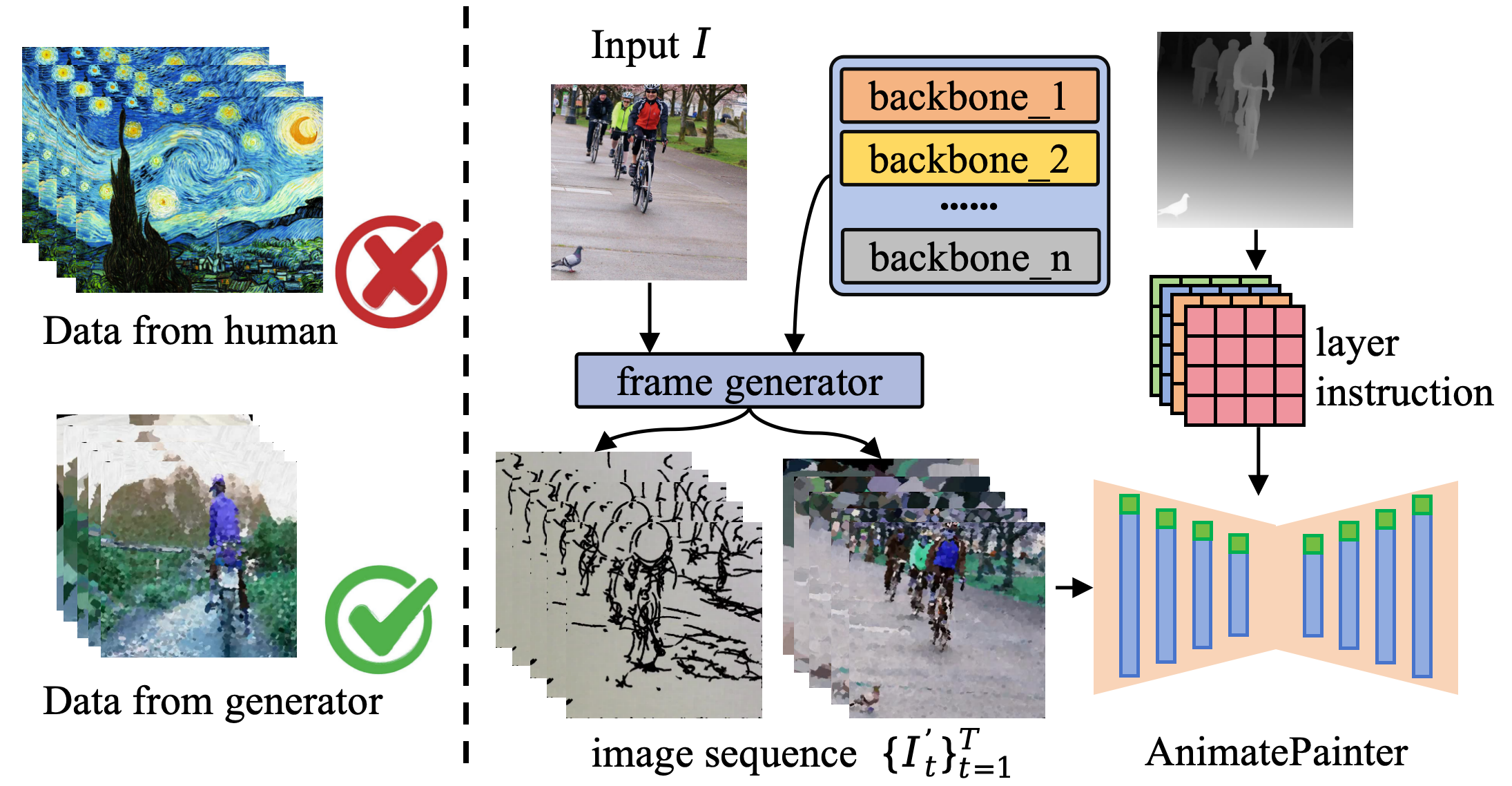}
    \caption{Our self-supervised method does not require expensive real painting process data, can quickly synthesize a large amount of painting data, and supports any SBR backbone.}
    \label{fig:first}
\end{figure}

To extend the task of image process generation to any type of image, we present \textbf{AnimatePainter}, a self-supervised rendering framework for reconstructing painting process. AnimatePainter consists of an self-supervised data generation method based on cognitive priors and a end-to-end process painting generator. It only requires easily accessible text-image pairs data to operate efficiently, without requiring actual painting data. Building upon a stroke-based renderer, AnimatePainter first reconstructs the painting process through cognitive painting principles that emulate human artistic decision-making, as shown on Fig. \ref{fig:first}. To address the domain gap between synthesized and authentic painting sequences, we subsequently introduce the DF-Encoder module, which injects geometric awareness through depth-map guided painting scheduling. A more detailed explanation of these two parts will be systematically elaborated in subsequent sections.

In particular, we define this task as a self-supervised image erasure problem that is not limited to a specific topic. Given an arbitrary image, we reconstruct the image by reverse erasing its content from complete image to blank canvas. Considering that human artists first conceptualize the overall framework and distribution of the target image when painting, and then confirm the priority order of each semantic part. Generally speaking, the distant background and the subjects of foreground are independent of each other, and the farther parts should be painted first. Inspired by previous manual rule-based works \cite{de2024segmentation,ganin2018synthesizing,PaintTransformer,LearningToPaint} and the painting skills of human artists, we provide some cognitive priors such as deep-layered-painting, contours-to-details, attention-area-limitation. 

Based on the above observations, we propose an end-to-end drawing process generation method that does not require test-time fine-tuning. Our method consists of two parts. In the first part, we use a self-supervised data generation method to generate a unified video dataset that includes the drawing process of each image. Only text-image pairs are required in this process, rather than actual drawing process videos. Our method is compatible with any existing strokes-based render methods, such as \cite{PaintTransformer,frida}. However, the real-simulate gap 
is inevitable. The discrepancy between the generated data distribution and the real data distribution prevents us from simply using the generated data directly for training, which can lead to overfitting of the model on "virtual data". To alleviate this issue, we introduce DF-Encoder to realize hierarchical painting control using a depth map generated by depth estimation. Like human artists typically divide the scene into foreground and background, we also utilizing the depth map to segment the entire scene into multiple painting sections. In the second part, a renderer based on a video generation diffusion model is proposed, which utilizes the depth hierarchical map and reference images to generate the canvas content for each frame. Different from previous approaches like Inverse Painting \cite{InversePainting}, we do not explicitly determine the painting area for each step, but instead use depth maps to guide the painting content in layers, as shown in Fig. \ref{fig:pipeline}.

We summarize our main contributions as follows:
\begin{itemize}
 \item We frame the process image generation task as a video generation problem, where intermediate step frames are generated between the first and last frames. To the best of our knowledge, AnimatePainter is the first to incorporate a video generation model into process image generation.
 \item  To overcome the challenge of obtaining real data, we propose a novel self-supervised data generation method that can efficiently produce large volumes of drawing process videos without the need for additional real drawing data. Our method is compatible with any stroke-based rendering technique and generates training data with a similar style.
 \item We introduce DF-Encoder to minimize the distribution gap between generated and real data, facilitating process image generation with a hierarchical structure.
\end{itemize}

\begin{figure}
    \centering
    \includegraphics[width=1\linewidth]{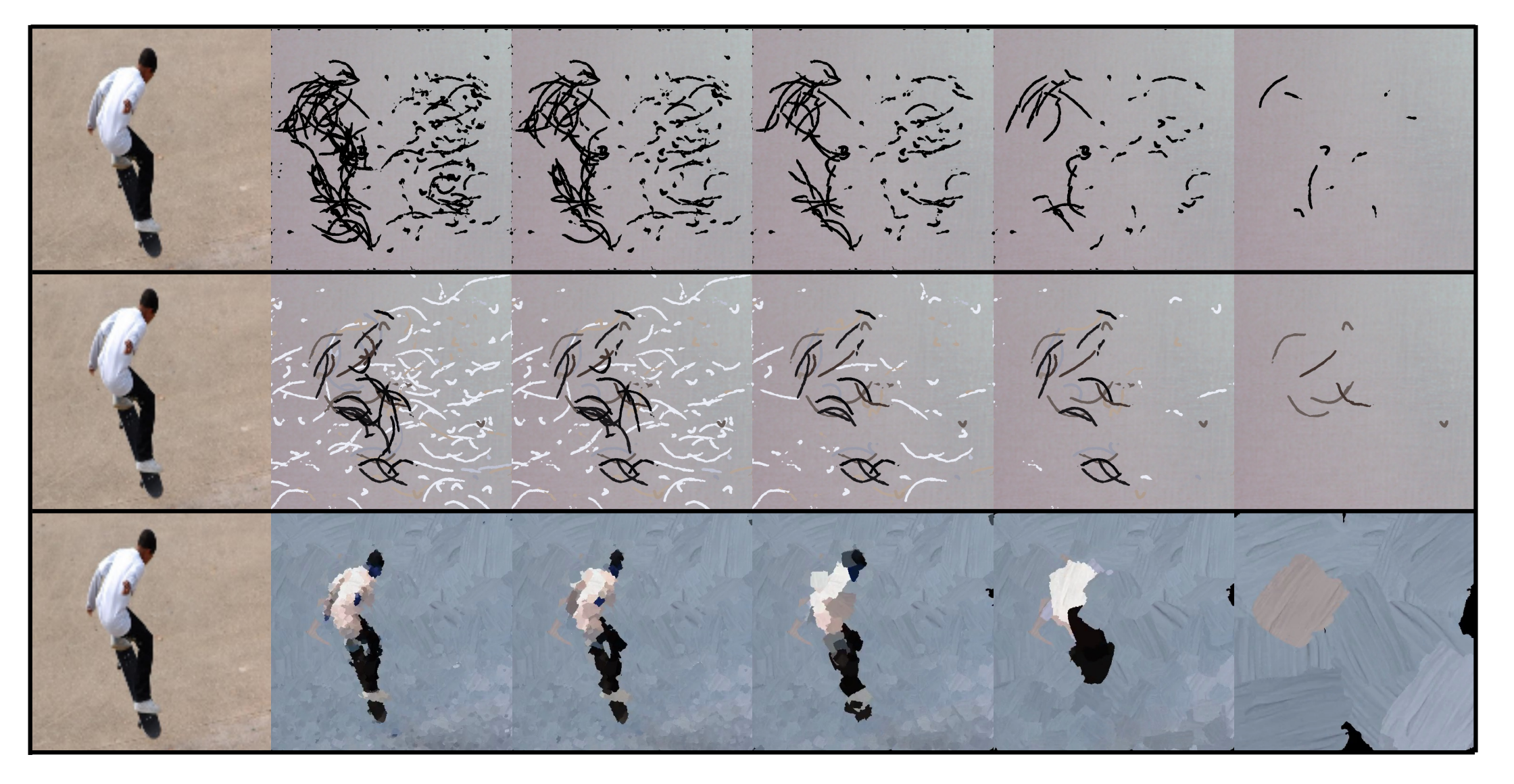}
    \caption{\textbf{Samples of dataset}. Some key frames of the generated data. The first two rows are based on Frida \cite{frida}(ink painting and colorful painting); the third row is based on PaintTransformer \cite{PaintTransformer}.}
    \label{fig:dataset}
\end{figure}

\section{Relation}
\label{sec:relation}
\subsection{Strokes-based Render}
Strokes-based Rendering, which employs line strokes to render realistic images, is widely used in robotic drawing \cite{frida,co-frida}, artistic creation \cite{PaintTransformer,InversePainting}, image stylization, and other fields. The painting process of rendering an image can be achieved by placing brush strokes. However, most current brush stroke-rendering methods \cite{frida,co-frida,neural_painters,ink_painting,CCD-BSM,clipdraw} focus on the modeling and image stylization of individual brush strokes, with only a small amount of work \cite{ink_painting,PaintTransformer} focusing on the generation of the drawing process.

In recent work, PaintTransformer \cite{PaintTransformer} predicts strokes sequences as tokens and employs a self-supervised random stroke generation method, enabling the generation of stunning images in the oil painting genre. To generate content step by step, some methods utilize reinforcement learning (RL) to constrain the content of strokes. Timecraft \cite{timecraft} proposes a probabilistic model to predict changes between two adjacent frames, achieving the synthesis of long videos; Inverse Painting \cite{ink_painting} employs a large language model to infer the next painting area, thereby generating a mask to guide the content generation at each step. Although they \cite{timecraft,ink_painting} have achieved impressive results, they are all limited to expensive real painting process datasets, making it difficult to expand in terms of painting styles.

The current methods exhibit some limitations that cannot be overlooked: (1) The approach relying on real painting datasets overly depends on manual painting processes. In many cases, such data is difficult to obtain and scarce, and it is challenging to extend to other painting forms (oil painting, ink painting, etc.); (2) Although many self-supervised and unsupervised methods have achieved success in predicting brushstrokes, the manual painting rules they rely on are not comprehensive and lack connection to the real world.

\begin{figure*}
    \centering
    \includegraphics[width=1\linewidth]{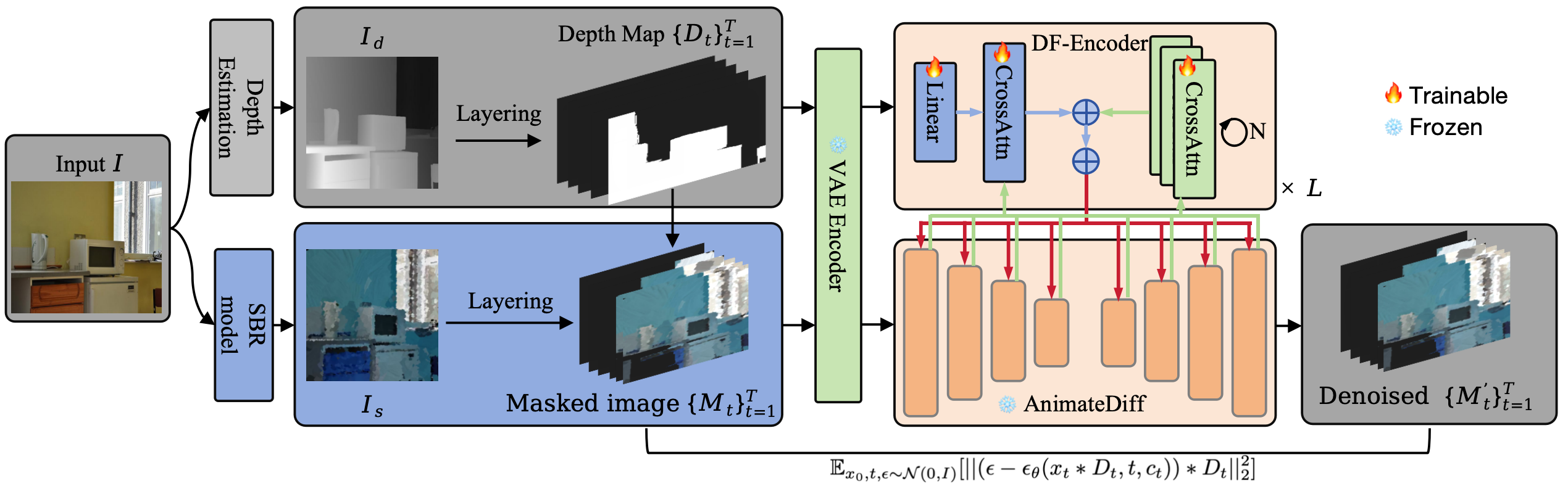}
    \caption{\textbf{Pipeline of AnimatePainter}. The input image $I$ will be used in two parts: (1)generating a depth map using a depth estimation model for hierarchical control. (2)rendering into strokes through the SBR model. In the layering stage, the depth map will be layered according to manual rules to obtain ${\{D_t\}}_{t=1}^T$. Similarly, it will also generate a layered image sequence ${\{M_t\}}_{t=1}^T$ based on the painting rules of human artists. After being encoded by VAE, ${D_t\}}_{t=1}^T$ enters the DF-Encoder to inject hierarchical control into the diffusion model. The entire painting process is generated in one step from end to end. And the strokes image $I_s$ is generated using PaintTransformer \cite{PaintTransformer} as backbone.}
    \label{fig:pipeline}
\end{figure*}

\subsection{Generative Diffusion Model}
\textbf{Text-to-image diffusion models.} \quad
In recent years, diffusion models \cite{denoising-diffusion-model, diffusion-beat-gan, stable-diffusion} have developed rapidly due to their excellent image generation capabilities. The emergence of latent diffusion models has made it possible to quickly generate high-quality images and has achieved significant commercial success. StableDiffusion \cite{stable-diffusion}, DALL-E2 \cite{DALL-E2}, Imagen \cite{Imagen} trained on large-scale image-text pair datasets, have become mainstream in text-to-image generation. DiTs \cite{DiTs} replaces the U-Net in the diffusion model with transformers, further improving the text-to-image generation performance and scalability. Since then, diffusion models based on DIT have become mainstream. Some cutting-edge text-to-image diffusion models like StableDiffusion3 \cite{stablediffusion3} have achieved even more impressive results by utilizing DIT \cite{DiTs} and flow matching \cite{flowmatching}. AnimateDiff \cite{guo2023animatediff,guo2023sparsectrl}
utilize motion module and control-net \cite{control-net}
adding motion dynamics to existing high-quality personalized T2Is and and enabling them to generate animations. In this paper, we adopt the AnimateDiff as our backbone to generate image sequences.

\noindent\textbf{Controllable generation.}\quad
In addition to using text prompts for conditional image generation, current diffusion models \cite{Animate_anyone,magicanimate,magicpose,dreambooth,lora} can add many additional inputs (e.g., edges, depth maps and segmentation maps) to achieve controllable image generation. Dreambooth \cite{dreambooth} and LoRa \cite{lora} are two mainstream controllable generation methods. They control the generation content of diffusion models through full-parameter fine-tuning and low-rank matrices fine-tuning, respectively. ControlNet \cite{control-net} utilize a trainable copy of the U-Net to integrate multimodal information. It freezes the original U-Net and only fine-tunes the trainable copy to apply control as a plugin. IP adapter \cite{ip-adapter} takes the reference images as feature inputs and achieves image-to-image generation by injecting cross-attention between the reference image and the denoising image into the source model. Other methods \cite{diffedit, noisecollage, dragdiffusion} use DDIM inversion \cite{DDIM} to generate noised images and realize image editing by manipulate the noised images.

\section{Method}
\label{sec:method}
Just like an artist drawing a complete picture step by step, unlike other existing methods, we define process image drawing as a video generation task. Specifically, we can complete two types of tasks: 1. Text2video: given a text describing the content of an image, we can generate a video about the painting process of the image based on the text content; 2. Image2video: given an image, using ControlNet \cite{control-net}, we can generate a video of the drawing process of this image. Now consider the image2video task. Given a target image $I$, we should reconstruct a T-frames video $\{I_1,I_2,......,I_{T-1},I_t\}$ , starting from a blank canvas to the target image or vice versa. 

Unlike traditional methods, we introduce a video generation model to achieve process image generation. This will make the generated video more coherent, and it is not limited to a certain type of painting.

\begin{equation}\label{other_loss}
 \mathcal{L} = Minimize(||I_{r}, I_{t}||^2).
\end{equation}

\subsection{Self-supervised data generation}
Previous works on strokes-based rendering focused on rendering real images into the form of strokes, which can be used in areas such as image style transfer, assisting people in learning painting, and robotic painting. Most of them employed optimization-based methods, solely aiming to ensure that the added strokes visually match the target image, that is, minimizing loss like equal \ref{other_loss}.

where $I_{r}$ and $I_{t}$ present the real image and the rendered image. Besides, the loss can be evaluated by MSE loss, MAE loss between images, or CLIP image similarity in high-level semantics. Moreover, most methods only consider the placement of each stroke and the modeling of strokes, with only a few focusing on the specific stroke order. However, simple and direct stroke placement cannot meet today's demands (education, robotic painting imitation, etc.). Unlike the stroke planning in calligraphy\cite{CCD-BSM}, discussing the stroke order in complex and colorfully varied stroke images is quite challenging. Based on observations of painting methods used by real artists and in modern art education, most works rely on manually constructed stroke rules to plan specific strokes, making it extremely easy to obtain training data. The most advanced current works collect a certain scale of real stroke data for training. Although these methods achieve the most impressive effects, they rely heavily on large-scale data that is difficult to obtain, and they are currently only effective on paintings such as landscape oil paintings, making it difficult to expand to other forms of content.

Based on the aforementioned discussion, we have abandoned expensive real data and adopted the traditional painting rule method, proposing a self-supervised process image dataset generation approach. Our method is a multimodal data construction approach that can quickly generate a large number of painting process videos for training, based on any previous strokes-based render methods and incorporating the painting rules of human artists.

Specifically, we first collected a large number of text-image pairs, which are relatively easy to obtain (e.g., LAION400M \cite{laion400m}). It includes real images and textual descriptions of their contents, used for training text-to-image diffusion models. To exclude low-quality data and ensure the method is not limited to certain high-quality text-image pair datasets, we also calculated the CLIP similarity between images and descriptive texts based on previous methods []. We reduced low-quality data by excluding images with low similarity. Then, we utilized the existing strokes-based render method to render real images into stroke forms. Next, we adopted different processing methods for two different strokes-based render methods. For the painting transformer method, which renders strokes frame by frame, our processing method is relatively simple. We directly sampled 12 frames of images (including 10 intermediate frames, 1 target frame, and 1 blank canvas frame) from the entire process as key strokes of the painting strokes. The number of frames is not fixed and can vary. For most other one-step rendering methods [ ], we divided the images into contour frames and detail frames based on the painting approach of human artists, who progress from overall contours to specific details. Specifically, we had previously obtained the optimized positions and rotation angles of each stroke. For each stroke in the target image, we calculated its score with all other strokes based on formula \ref{eq:sort_distance} to evaluate the density of that line. We utilized a clustering method akin to k-nearest neighbors. Consider the equal \ref{eq:sort_distance}, we rendered the target image into $t$ strokes and select one of the strokes $S_i\in\{S_1,......,S_t\}$ to calculate score with each other.$\Psi(\cdot,\cdot)$ means euclidean distance, $W$ is the width of target image and $r_i$ is the absolute rotation angle of the strokes.

\begin{equation}\label{eq:sort_distance}
 Score_i = \sum_{k=1}^{n} (\Psi(S_i,S_k)<0.1*W \:\: \& \:\:|r_i-r_k|< \frac{\pi}{4}.
\end{equation}

Then we sort all the strokes based on their scores, which indicate the density of their locations. Following the painting rule of drawing the outline first and then adding details, we continuously remove strokes from dense areas in the complete image. We divide the process from the target image to a blank canvas into 10 steps. Using this method, we can obtain key frames that adhere to the painting rules.

Table. \ref{number_of_dataset} demonstrates a comparison between the amount of data generated by our method and methods \cite{InversePainting} based on real data. We can directly generate a large amount of self-supervised data for training, which far exceeds the amount of data generated by existing methods.

\begin{table}
  \caption{\textbf{Datasets}. The amount of data.}
  \label{number_of_dataset}
  \centering
  \begin{tabular}{lc}
    \toprule
    Method     & Number of videos  \\
    \midrule
    Inverse painting \cite{InversePainting} & 294 \\
    \midrule
    Ours(based on Frida \cite{frida}) & 1046  \\
    Ours(based on Frida \cite{frida} ink)  & 1070\\
    
    Ours(based on PaintTransformer \cite{PaintTransformer}) & 20000 \\
    \bottomrule
  \end{tabular}
\end{table}

\begin{figure}
    \centering
    \includegraphics[width=1\linewidth]{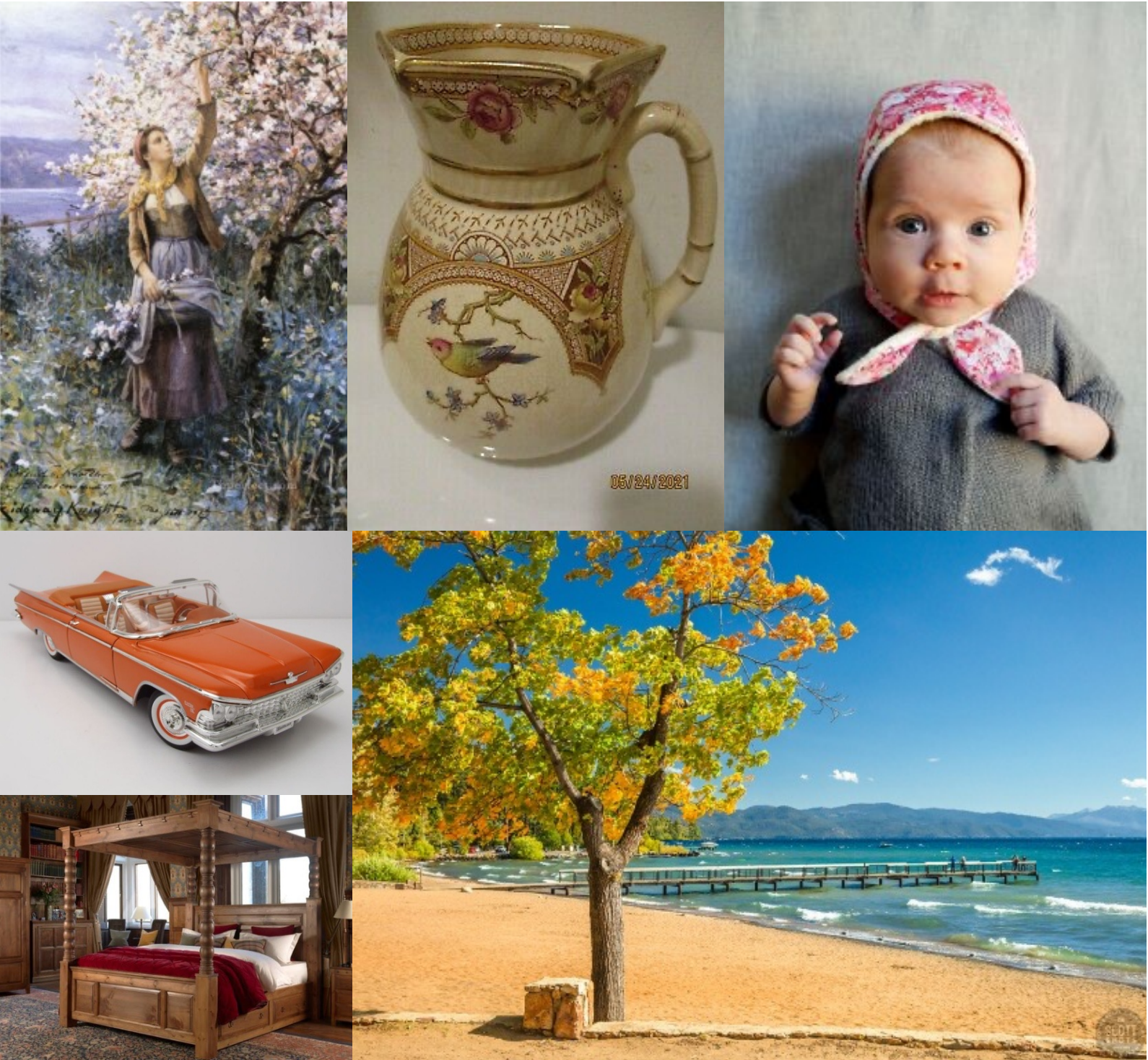}
    \caption{\textbf{Samples of evaluation dataset}. Including oil paintings, landscape images, portraits, and more.}
    \label{fig:data_sample}
\end{figure}

\begin{figure*}
    \centering
    \includegraphics[width=1\linewidth]{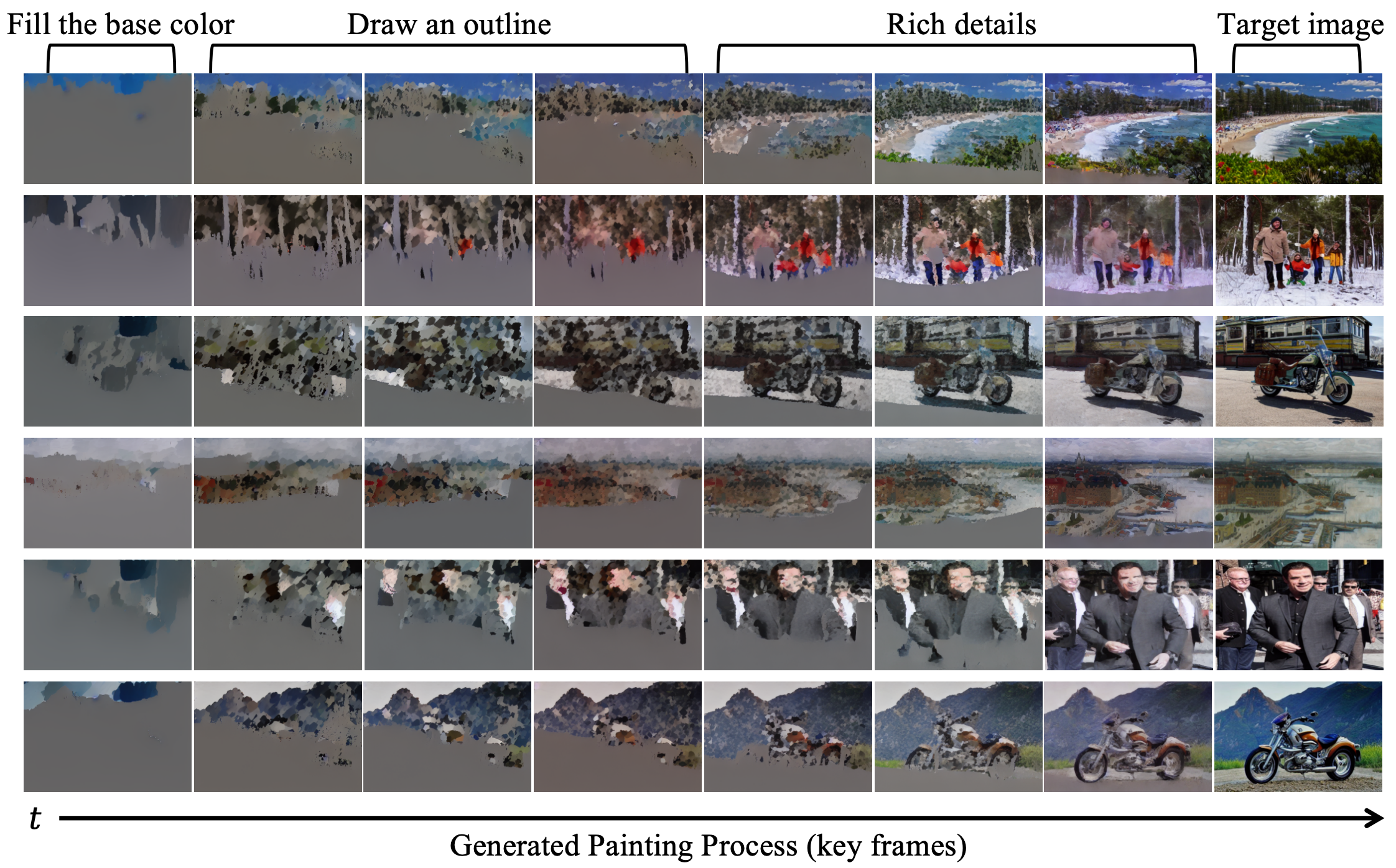}
    \caption{\textbf{Generated Painting Process}. At the beginning of painting, AnimatePainter fills in the basic background color and then focuses on depicting the rough outline of the object, generating only relatively blurry content at this stage. Subsequently, the model begins to refine the details and enhance the sharpness of the painting.  Meanwhile, 
    the model adds details from far to near during the painting process, which aligns with our previous discussion.}
    \label{fig:generated_pp}
\end{figure*}

\subsection{End-to-end process painting generation}
Considering the inefficient optimization and lack of diversity in traditional methods, and to address the poor flexibility of the aforementioned strokes-based methods, we have introduced a video generation diffusion model to achieve highly consistent procedural image generation. We incorporate the depth map of the target image (generated by a depth estimation model) as an additional control condition into the video diffusion model, using hierarchical depth information to guide the generation of procedural videos. To our knowledge, we are the first to introduce a video generation model into procedural image generation.

Following the Stable diffusion model \cite{stable-diffusion,guo2023animatediff}, we also utilize a latent denoising loss $\mathcal{L}_{LDM}$ to Minimize the KL divergence between the generated image distribution and the target image distribution, as demonstrated in Eq. \ref{L_LDM}:

\begin{equation}\label{L_LDM}
 \mathcal{L}_{LDM} = \mathbb{E}_{x_0, t, \epsilon \sim \mathcal{N}(0,I)}[||\epsilon - \epsilon_\theta(x_t, t, c_t)||^{2}_{2}].
\end{equation}

\noindent \textbf{Layering the target image.} In our method, the target image is divided into T (we take 10) frames of key frames. To enable the model to mimic the painting process of human artists and generate process videos with different levels of foreground and background, we stratify the target image according to the estimated depth map. At the same time, considering that the depth range of the depth map generated by depth estimation mostly follows a normal distribution, using a segmentation method with a fixed threshold can lead to significant differences in the number of pixels between different layers. Therefore, we determine the segmentation threshold based on the number of pixels, ensuring that the number of pixels contained in each layer of the image is basically consistent. Previous work \cite{frida,co-frida,InversePainting} has used semantic segmentation and other methods to segment the target image to enhance the model's generation performance. However, according to our observation, simply removing semantic objects from the image can lead to discontinuity in the image and make it no longer consistent with the text description. Most importantly, using semantic segmentation directly determines that the model draws content based on semantic objects, which is contrary to the painting style of human artists. For humans, drawing isolated individual objects can lead to structural inconsistency in the entire picture, and most people will first form the overall composition and then refine the content.

\noindent \textbf{Multimodal depth information fusion.}
There are various commonly used methods for integrating multimodal information into models, such as add, concat, attention, and so on. Our approach is based on AnimateDiff \cite{guo2023animatediff} and sparsectrl \cite{guo2023sparsectrl}, which is a controllable plug-and-play module turning most community text-to-image models into animation generators. The core of AnimateDiff is a plug-and-play motion module that learns reasonable motion priors from video datasets. It's able to capture and learn the motion relationships between frames, ensuring smoother transitions between frames when generating videos. 
From another perspective, we regard the task of generating procedural images as a video generation task based on prior knowledge. Similar to generating moving images, our goal is to generate the drawing process while maintaining the overall stillness of the image. According to our observation, the main function of the motion module in animatediff is to achieve smooth transitions in frame generation, which is also consistent with generating the transition process between two frames of painting. Therefore, we can transfer it to our task.

However, directly using the motion module can lead to slight movements in the generated images and fail to generate intermediate images effectively. To solve this problem, as shown in Fig \ref{fig:pipeline}, we propose \textbf{DF-Encoder}(depth fusion encoder), a depth fused cross-attention module which combine with each motion module to introduce hierarchical depth information, helps guide the model to generate results that conform to hierarchy. Our method achieves efficient depth hierarchy control with only a small number of additional parameters, and requires no modifications to the source model.

Assuming the U-Net model is $\mathcal{F}(x;\theta)$, our multimodal depth information fusion can be represented as below:
\begin{equation}\label{eq5}
 Q=W^Qz_i, K=W^K\Psi(f_d), V=W^V\Psi(f_d),
\end{equation}

\begin{equation}\label{eq6}
 z_{out} = Softmax(\frac{QK^{T}}{\sqrt{d}})V,
\end{equation}

\noindent where $\Psi(·)$ is MLP, $\{z_i\}_{i=0}^{N}$ is the n-th hidden state in Unet and $f_d$ is the latent state of layered depth map. As we get the depth fusion feature $z_{out}$, we can inject it into the Motion Module \cite{guo2023animatediff}.

\begin{equation}\label{eq:noise}
 \epsilon_{t,\theta} = \mathcal{F}(\varepsilon_t, x_t*D_t, c, t, \theta),
\end{equation}

\begin{equation}\label{eq:loss}
 \mathcal{L}_{layer} = \mathbb{E}_{x_0, t, \epsilon \sim \mathcal{N}(0,I)}[||(\epsilon - \epsilon_\theta(x_t*D_t, t, c_t))*D_t||^{2}_{2}].
\end{equation}

After that, our prediction formula for noise in the image denoising process is shown in Eq. \ref{eq:noise}. $D_t$ is the layered masks which is shown in Fig. \ref{fig:pipeline}. Our final loss $L_layer$ can be presented as Eq. \ref{eq:loss}, $(*)$ means the element-wise multiplication of a matrix.


\section{Experiments}
\label{sec:experence}

%
%

\subsection{Training detail and dataset}
We use different SBR methods \cite{frida,PaintTransformer} as the backbone to generate training data from COCO-2017 \cite{coco_dataset} dataset. For methods capable of generating intermediate steps, we sample T frames at fixed intervals. For methods that only perform stroke rendering, we adopt the method mentioned in section \ref{sec:method} for sampling, generating training videos of T frames. The dataset size generated by our method during training are shown in Table \ref{number_of_dataset}.

For our method, we employed fixed settings and random number seeds to minimize randomness. We trained on the dataset for 25,000 steps with a learning rate of 1e-4 and a batch size of 1. Our training can be completed in less than 10 hours on a single NVIDIA A6000. For other methods used for comparison, we adopted their official code and used the pre-trained models provided by them along with their default config.

\begin{table*}
  \caption{\textbf{Comparison with baselines}. AnimatePainter(ours) is superior to existing self-supervised and fully supervised methods, with only a slight disadvantage over fully supervised methods in terms of SSIM.}
  \label{metric}
  \centering
  \begin{tabular}{lcccccccc}
    \toprule
    Method & Supervision & Painting type & CLIP-I$\uparrow$ & DINOv2$\uparrow$  & FID$\downarrow$ & LPIPS$\downarrow$ & SSIM$\uparrow$ & DDC$\downarrow$\\
    
    \midrule
    Frida(ink) \cite{frida} & Self & Ink Strokes & \underline{0.662} & 0.257 & 395.665 & 0.682 & 0.194 & - \\
    Frida(colorful) \cite{frida} & Self & Colorful Strokes & 0.659 & 0.189 & 321.778 & 0.592 & 0.207 & - \\
    PaintTransformer \cite{PaintTransformer} &Self & Oil Painting & 0.528 & 0.470 & 150.627 & 0.342 & 0.518 &4.090 \\
    Timecraft \cite{timecraft}& Full & Oil Painting & 0.330 & \underline{0.560} & 131.968 & 0.637 & \textbf{0.550} &9.696\\
    Inverse painting \cite{InversePainting}& Full & Oil Painting & 0.426 & 0.114 & 158.860 & 0.707 & 0.343 & 13.807 \\
    \midrule
    $Ours\text{-}DF$ & Self & All Painting& 0.090 & 0.284 & 170.170 & 0.456 & 0.332 &1.852 \\
    $Ours\text{-}DL$ & Self & All Painting& 0.530 & 0.534 & \underline{87.684} & \underline{0.311} & 0.481 & \underline{1.256} \\
    ${Ours}_{all}$ & Self & All Painting& \textbf{0.698} & \textbf{0.622} & \textbf{66.235} & \textbf{0.301} & \underline{0.524} &\textbf{1.238}\\
    \bottomrule
  \end{tabular}
\end{table*}

\subsection{Metrics}
\subsubsection{Evaluation dataset}
To evaluate and compare performance, we randomly sampled 1000 images from the LAION-art dataset, a subset of LAION \cite{laion400m} where the image aesthetic score of each image exceeds 8, as target images. It encompasses various types of images at different resolutions, including oil paintings, landscape images, portraits, and more. Some examples are shown in the Fig. \ref{fig:data_sample}.

\subsubsection{Evaluation metrics}
For process image generation, there are currently no particularly comprehensive evaluation metrics. Previous work \cite{PaintTransformer,robotPaint1,frida,co-frida,InversePainting,dreambooth,stable-diffusion,ganin2018synthesizing,LearningToPaint,de2024segmentation} has proposed some commonly used metrics in the field of image generation, such as FID \cite{FID} and CLIP \cite{CLIP} score. We also adopted the following evaluation metrics based on the evaluation method of InversePainting \cite{InversePainting}:
\begin{itemize}
 \item \textbf{CLIP/DINO2 score.} We utilize the CLIP \cite{CLIP} and DINOv2 \cite{DINO} multimodal model to assess the similarity between the test image and the final generated complete image, thereby evaluating the fidelity of the method to the target image. This approach is also employed in strokes-based rendering to evaluate the capability of generating renderings.

 \item \textbf{FID \cite{FID}/LPIPS \cite{lpips}/SSIM \cite{ssim} score.} They calculated the aesthetic index, perceptual similarity, and structural similarity between the generated image and the target image to comprehensively evaluate the quality of the generated image.

  \item \textbf{DDC \cite{InversePainting} score.} It calculates the LPIPS score between the key frame sequence and the target image, and the corresponding curve was plotted. The difference between this curve and the theoretical curve was then calculated using the DTW(Dynamic Time Warping) \cite{DTW} algorithm. Reflecting the dynamics of an actual painting process.

\end{itemize}

\begin{figure}
    \centering
    \includegraphics[width=1\linewidth]{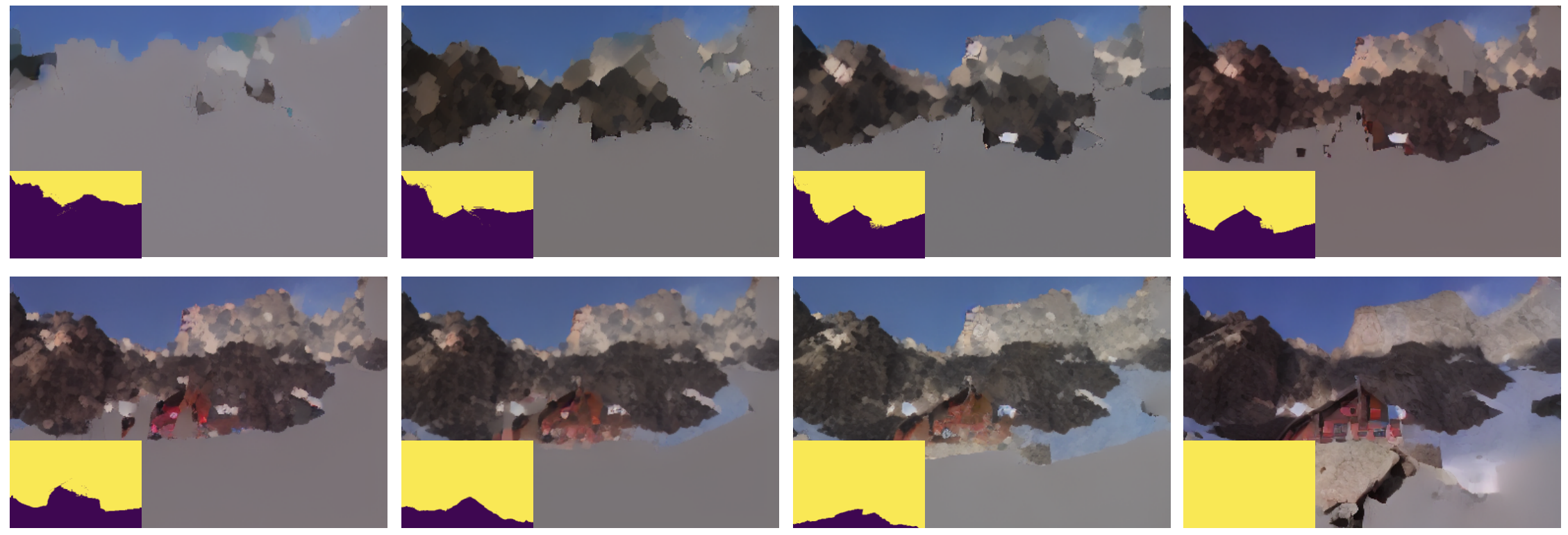}
    \caption{\textbf{Qualitative results with multimodal depth fusion}. The yellow section represents the guidance for the model's inference region for the next frame. We utilize this non-mandatory mask to guide the generation process.}
    \label{fig:depth_guiden}
\end{figure}

\begin{figure*}
    \centering
    \includegraphics[width=1\linewidth]{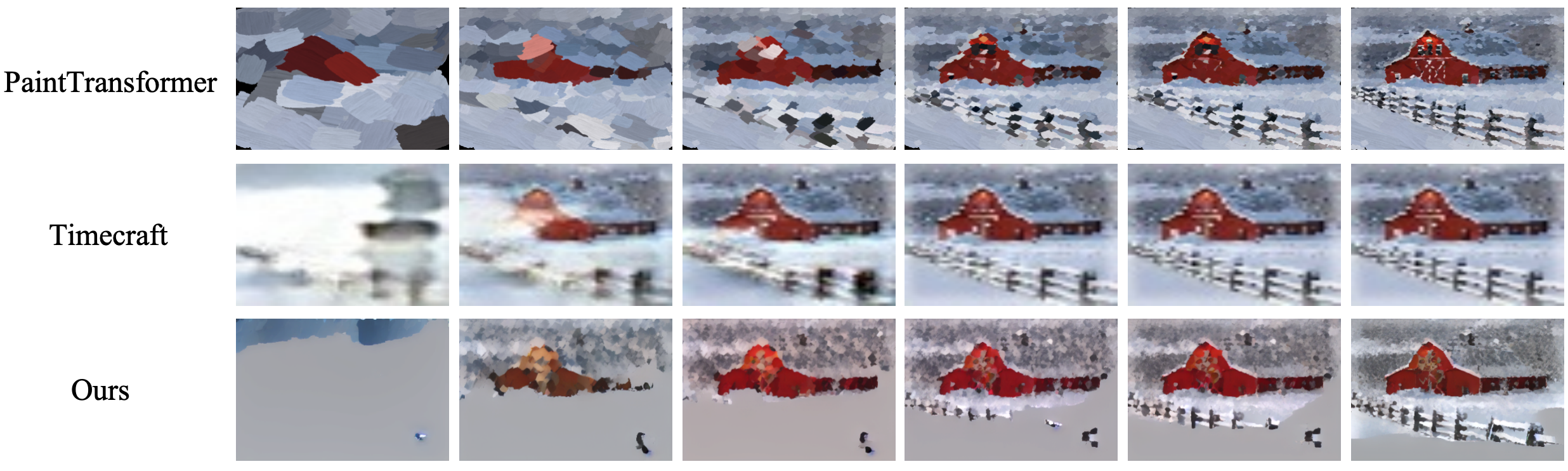}
    \caption{\textbf{Qualitative comparison}. We uniformly sampled key frames from images generated by various methods. On the generated painting sequence, although PaintTransformer has good oil painting effects, it merely optimizes brushstrokes and cannot reflect the real painting process. TimeCraft can draw images by region, but its 50x50 resolution makes most of the generated images very blurry.  Our method can rely on deep hierarchical control to draw background and foreground images, achieving human-like painting effects.}
    \label{fig:compare}
\end{figure*}

\subsubsection{Qualitative results}
Fig.\ref{fig:generated_pp} shows our qualitative generation results.  It can be observed that in the initial stage of generation, AnimatePainter applies a large brush stroke to paint the overall background color (first column). Subsequently, the model attempts to sketch the outline of the object (columns 2-4). Due to our predefined painting rules for human artists, which transition from rough outlines to specific details, the generated image content is relatively blurry at this point, forming the layout and shape of the object. Then, the model begins to add details to various parts (columns 5-7), refining all blurry parts to gradually produce clearer facial features, textures, etc. Meanwhile, 
the model fills in foreground details from far to near until all parts are fully painted.
Fig.\ref{fig:generated_pp} shows our qualitative generation results.  It can be observed that in the initial stage of generation, AnimatePainter applies a large brush stroke to paint the overall background color (first column). Subsequently, the model attempts to sketch the outline of the object (second and third columns). Due to our predefined painting rules for human artists, which transition from rough outlines to specific details, the generated image content is relatively blurry at this point, forming the layout and shape of the object. Then, the model begins to add details to various parts (columns 3-5), refining all blurry parts to gradually produce clearer facial features, textures, etc. Meanwhile, as can be seen from the red rectangular portion, the model fills in foreground details from far to near until all parts are fully painted.

Fig. \ref{fig:depth_guiden} illustrates the deep hierarchical fusion achieved through the DF-Encoder during the generation process. The yellow section represents the guidance for the model's inference region for the next frame.

\subsubsection{Comparison with baseline}
We conducted a detailed comparison with the baselines at both qualitative and quantitative levels. In terms of qualitative comparison, as shown in Figure \ref{fig:compare}, we uniformly sampled the same key frames from the generated image sequence for comparison. The PaintTransformer is more akin to stylizing images, failing to reflect the connection between the painting process and the real world. The low-resolution images produced by timecraft exhibit numerous artifacts, compromising the generation quality. Conversely, our method employs deep hierarchical fusion to guide the model in drawing images from background to foreground, aligning with the intuitive painting techniques of human artists and achieving remarkable results.

In terms of quantitative comparison, as illustrated in Table. \ref{metric}, our method outperforms all current self-supervised methods in terms of indicators, and also substantially surpasses fully supervised methods.

\subsection{Ablation study}
Our ablation study consists of two parts. Firstly, we will evaluates the different components in the pipeline using 7 variants. (1)${Ours}_{all}$ the full model includes input and depth map layering, DF-Encoder. (2)$Ours\text{-}DF$ the full model without DF-Encoder. (3)$Ours\text{-}DL$ the full model without DF-Encoder and input layering. As shown on Table.\ref{metric}, our complete model exhibits the best performance, and removing any of its components will affect the quality of the generated images.

\begin{figure}
    \centering
    \includegraphics[width=1\linewidth]{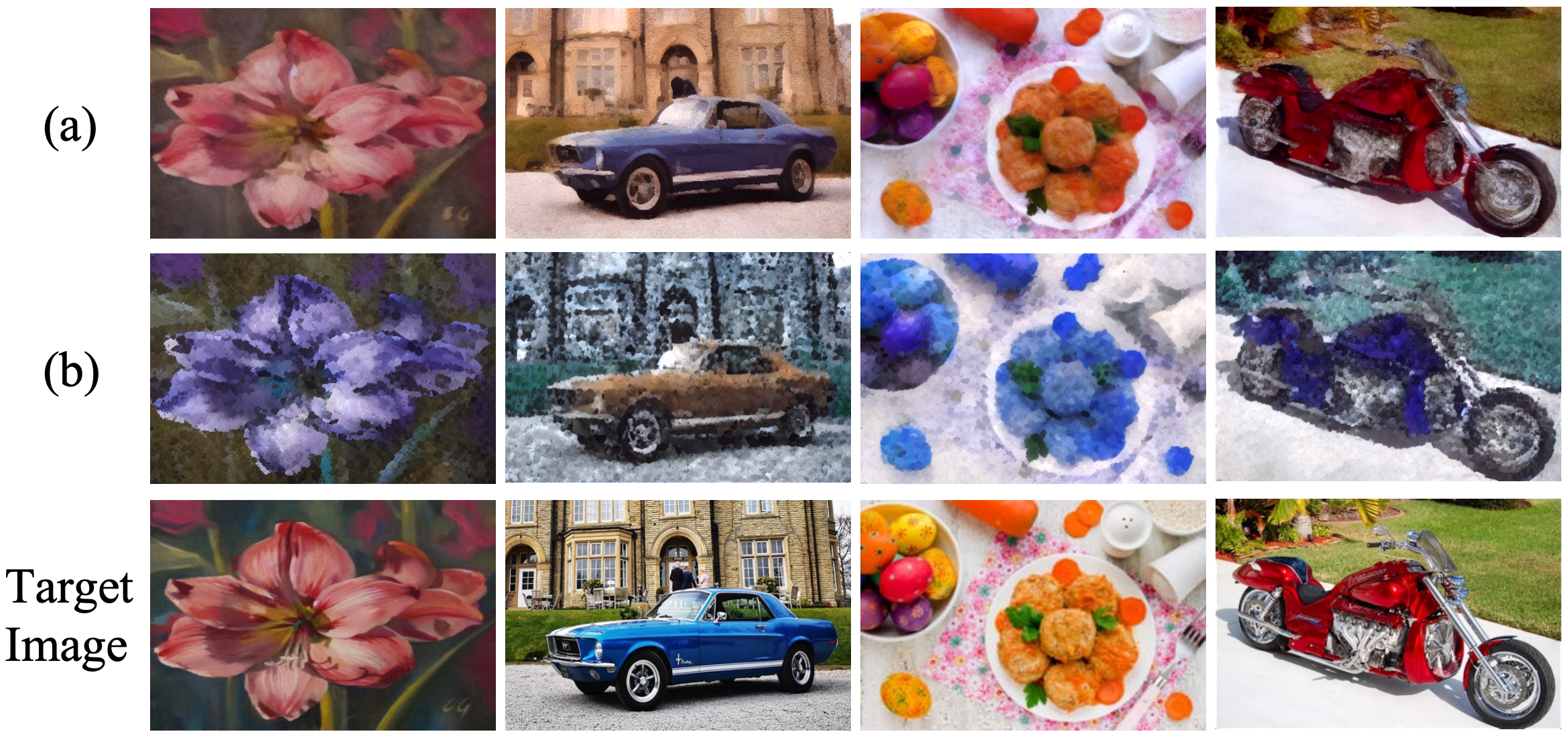}
    \caption{\textbf{Different Control Strategies}. Due to the distribution deviation between the images generated by the SBR model and the target image, the generated images may deviate from the target image (as shown in (b)). Therefore, we use stroke images as controls during training (although this may result in inconsistent control conditions between the training and inference stages), which can effectively eliminate distribution differences. As shown in (a).}
    \label{ablation1}
\end{figure}

Another part of the experiment verifies the robustness of our method to input bias, as shown on Fig.\ref{ablation1}. We utilize consistency control to observe the impact of the SBR model used on the generated results. (a) Using strokes image as control image. (b) Using target image as control image.

\section{Conclusion}

In this paper, we propose AnimateDiff, a self-supervised method for reconstructing the painting process that can generate process images without requiring any real data. Our method follows the painting rules of human painters and performs stroke erasure on the images generated by existing SBR methods to generate training data. At the same time, we proposed DF-Encoder to alleviate the bias of input data and bridge the gap between generated images and real world.

\subsection{Limitation}

As an end-to-end generation model, AnimatePainter does not explicitly distinguish between individual steps, which to some extent leads to ambiguity in our generation process and makes it impossible to specify the generation content for each step. Constrained by the performance of video generation models, our method has a relatively strict frame limit, making it unable to generate long videos with extended durations.
{
    \small
    \bibliographystyle{ieeenat_fullname}
    \bibliography{main}
}

\end{document}